\documentclass[10pt,twocolumn,letterpaper]{article}

\usepackage{ragged2e} 
\usepackage{booktabs,makecell, multirow, tabularx}
\usepackage{fontawesome}
\usepackage[preprint]{cvpr}      

\definecolor{cvprblue}{rgb}{0.21,0.49,0.74}
\usepackage[pagebackref,breaklinks,colorlinks,allcolors=cvprblue]{hyperref}

\def\paperID{***} 
\def\confName{CVPR}
\def\confYear{2025}

\title{GCUNet: A GNN-Based Contextual Learning Network for Tertiary Lymphoid Structure Semantic Segmentation in Whole Slide Image}
\author{
    Lei Su\textsuperscript{1,2},
    Zonghao Liu\textsuperscript{3}, Junxian Wu\textsuperscript{4}, 
    Zewen Sun\textsuperscript{1,2}, 
    Jiaxuan Wen\textsuperscript{1,2},
    Lu Zhu\textsuperscript{1,2},\\
    Xuqing Geng\textsuperscript{1,2},
    Tianwang Xun\textsuperscript{1,2}, 
    Jing Kuang\textsuperscript{5},
    Lizhi Shao\textsuperscript{6}, 
    Yang Du\textsuperscript{1,2}\\ 
    \textsuperscript{1} CASMI, Institute of Automation, Chinese Academy of Sciences (CASIA)\\
    \textsuperscript{2} School of Artificial Intelligence, University of Chinese Academy of Sciences (UCAS)\\
    \textsuperscript{3}Clinical Oncology School of Fujian Medical University, Fujian Cancer Hospital\\ 
    \textsuperscript{4} School of Computer Science and Engineering, Southeast University\\
    \textsuperscript{5} Institute of Pathology, Tongji Hospital, Tongji Medical College,\\ Huazhong University of Science and Technology\\
    \textsuperscript{6} School of Internet, Anhui University\\
    \texttt{\{sulei2023, sunzewen2022, wenjiaxuan2023, zhulu2024\}@ia.ac.cn} \\
    \texttt{\{gengxuqing2024, xuntianwang2022, yang.du\}@ia.ac.cn}\\
    \texttt{liuzonghao42@gmail.com},
    \texttt{junxianwu@seu.edu.cn},\\  \texttt{kuangjing@tjh.tjmu.edu.cn}, \texttt{24115@ahu.edu.cn}
}

\begin{document}
\maketitle
\begin{abstract}
We focus on tertiary lymphoid structure (TLS) semantic segmentation in whole slide image (WSI). Unlike TLS binary segmentation, TLS semantic segmentation identifies boundaries and maturity, which requires integrating contextual information to discover discriminative features. 
Due to the extensive scale of WSI (\textit{e.g.}, 100,000 $\times$ 100,000 pixels), the segmentation of TLS is usually carried out through a patch-based strategy.
However, this prevents the model from accessing information outside of the patches, limiting the performance.
To address this issue, we propose GCUNet, a GNN-based contextual learning network for TLS semantic segmentation.
Given an image patch (target) to be segmented, GCUNet first progressively aggregates long-range and fine-grained context outside the target.
Then, a Detail and Context Fusion block (DCFusion) is designed to integrate the context and detail of the target to predict the segmentation mask.
We build four TLS semantic segmentation datasets, called TCGA-COAD, TCGA-LUSC, TCGA-BLCA and INHOUSE-PAAD, and make the former three datasets (comprising 826 WSIs and 15,276 TLSs) publicly available to promote the TLS semantic segmentation \footnote{We will release the datasets after the acceptance of this paper.}.
Experiments on these datasets demonstrate the superiority of GCUNet, achieving at least 7.41\% improvement in mF1 compared with SOTA.
\end{abstract}    
\section{Introduction}
\label{sec:intro}

\begin{figure}[t]
    \centering
    \begin{subfigure}{0.9\linewidth}
        \includegraphics[width=\linewidth]{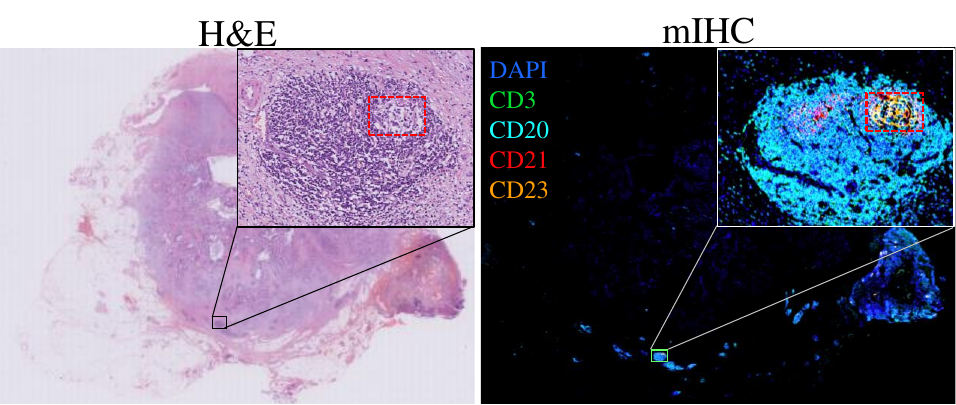}
        \caption{Comparison of SEL-TLS in H\&E and mIHC.
        The SEL-TLS contains a germinal center (GC), which is highlighted by the red box. In H\&E, a pale staining region represents GC. In mIHC, the GC is identified by CD23$^+$.
        The mIHC includes DAPI (deep blue, for nucleus), CD3 (green, for T cells), CD20 (light blue, for B cells), CD21 (red, for follicular dendritic cells), and CD23 (orange, for GC).
        }
        \label{fig:1_a}
    \end{subfigure}
    
    
    \begin{subfigure}{0.9\linewidth}
        \includegraphics[width=\linewidth]{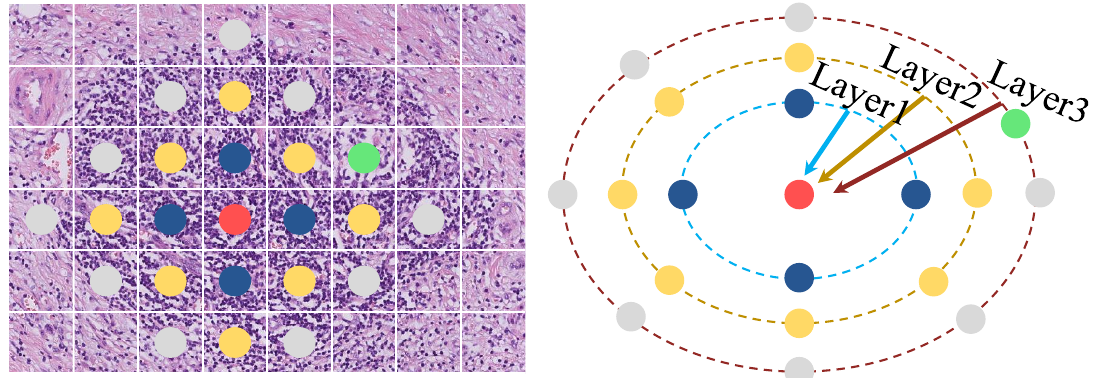}
        \caption{\textbf{Left.} The SEL-TLS is divided into multiple patches for segmentation. The red node represents the target patch for segmentation, the blue nodes are first-order neighbors, the yellow nodes are second-order neighbors, and the gray and green (GC, which determines TLS maturity) nodes are third-order neighbors. \textbf{Right.} GCUNet progressively aggregates contextual information outside the target patch by GCN layers.}
        \label{fig:1_b}
    \end{subfigure}
    \caption{GCUNet gathers discriminative features by aggregating contextual information outside the target patch.}
    \vspace{-10pt}  

    \label{fig:1}
\end{figure}

Tertiary lymphoid structure (TLS) is an aggregate of immune cells that can be classified into three levels of maturity: early TLS (E-TLS), primary follicular-like TLS (PET-TLS) and secondary follicle-like TLS (SEL-TLS) \cite{42, tls_intro_three_stage}.
In most solid tumors, the presence of TLS is closely associated with the anti-tumor immune response, which is significantly influenced by the maturity of TLS.
To identify the maturity levels of TLS, 
multiplex-immunohistochemistry (mIHC) is commonly used to detect specific molecular expressions such as CD21$^+$ for PET-TLS and CD23$^+$ for SEL-TLS.
However, the widespread adoption of this approach is limited by time and economic costs, as well as available examination techniques.
Fortunately, the molecular expressions also lead to morphological changes in nucleus and tissue structures \cite{7,8}.
As shown in \Cref{fig:1_a}, SEF-TLS not only expresses CD23$^+$ but also exhibits germinal center (GC) in whole slide image (WSI) stained with hematoxylin and eosin (H\&E).
 Given this, the identification of TLS in WSI is important in tumor diagnosis and treatment.



In recent years, computational pathology (CPath) \cite{23} has attracted increasing attention for its wide range of applications such as cancer classification \cite{10,11,12}, tumor grading \cite{13,14}, survival analysis \cite{15,17}, and biomarker prediction \cite{18,19, contextual_learning_cell_HER2_pre_2022_MIA}.
Binary segmentation is a critical approach for delineating 
the boundaries of TLS \cite{48,49,57,multi_sacle_cnn_04_pixel_level, multi_sacle_cnn_05_pixel_level}. 
Due to the extensive scale of WSI (\textit{e.g.}, 100,000 $\times$ 100,000 pixels), the segmentation procedure is often carried out in two steps:
First, the WSI is divided into numerous image patches, and each patch is processed by a segmentation model.
Second, the patch-level results are assembled into the entire TLS segmentation image.
However, for the TLS semantic segmentation task, 
this approach lacks awareness of contextual information beyond the target patch, which restricts its ability to uncover discriminative features and thereby limits segmentation performance.


In WSI analysis, contextual learning methods based on CNNs \cite{multi_sacle_cnn_01_pixel_level, 
multi_sacle_cnn_02_pixel_level, multi_sacle_cnn_03_pixel_level, multi_sacle_cnn_04_pixel_level,
multi_sacle_cnn_05_pixel_level,
multi_sacle_cnn_01_path_level},
Transformers \cite{multi_sacle_trans_06_wsi_level_wsi_level, multi_sacle_trans_07_patch_level, multi_sacle_trans_image_04_wsi_level}, and GNNs \cite{multi_sacle_graph_01_wsi_level, multi_sacle_graph_03_patch_level, multi_sacle_graph_05_wsi_level, multi_sacle_graph_08_patch_level, multi_sacle_graph_09_wsi_level, contextual_learning_cell_HER2_pre_2022_MIA} have been developed to capture multi-scale contextual information or enhance global context awareness.
However, much of the prior work has primarily focused on WSI-level tasks \cite {multi_sacle_graph_01_wsi_level, multi_sacle_graph_05_wsi_level, multi_sacle_graph_09_wsi_level, multi_sacle_trans_06_wsi_level_wsi_level} and patch-level tasks \cite{multi_sacle_cnn_01_path_level, multi_sacle_graph_03_patch_level, multi_sacle_graph_08_patch_level, multi_sacle_trans_07_patch_level}, such as survival risk prediction and patch classification.
Existing methods for pixel-level tasks
\cite{multi_sacle_cnn_01_pixel_level, multi_sacle_cnn_02_pixel_level,multi_sacle_cnn_03_pixel_level,multi_sacle_cnn_04_pixel_level,multi_sacle_cnn_05_pixel_level}
rely exclusively on CNNs and low-resolution images, which limits their ability to capture long-range and fine-grained contextual information.
Applying contextual learning to pixel-level segmentation tasks in WSI, such as TLS semantic segmentation, remains an area worthy of further exploration.

To address the issue, we propose a GNN-based contextual learning network (GCUNet) to capture long-range contextual and fine-grained outside target patch (target) for TLS semantic segmentation.
As illustrated in \Cref{fig:1_b}, this model progressively aggregates contextual information outside the target.
Additionally, a Detail and Context Fusion
block (DCFusion) is designed to perform a semantic-level fusion of the contextual and detailed information.
We build four cancer-type TLS semantic segmentation datasets and demonstrate the superiority of GCUNet, achieving at least a 7.41\% improvement in mF1 over SOTA.
The main contributions of our method include:
\begin{itemize}
    \item We focus on a new task, i.e., TLS semantic segmentation in WSI. To the best of our knowledge, we are the first to capture contextual information outside of the target patch for TLS semantic segmentation.
    \item We present a new GNN-based contextual learning method GCUNet for TLS semantic segmentation. GCUNet leverages GCNs to flexibly aggregate long-range and fine-grained contextual information outside patches, while the designed DCFusion performs a semantic-level fusion of detailed and contextual information to predict segmentation masks.
    
    \item We gather four datasets from different cancer types for validation. Considering the difficulty of acquiring pixel-level annotations in WSI, we release three annotated datasets based on TCGA (TCGA-COAD, TCGA-LUSC, TCGA-BLCA, comprising 826 WSIs and 15,276 TLSs) to promote the TLS semantic segmentation.
\end{itemize}

\section{Related work}
\label{sec:relate_work}

\subsection{TLS Segmentation in WSI}

Three maturity stages of TLS are classified based on the presence of follicular dendritic cells or GC \cite{1}.
The existing end-to-end methods for TLS segmentation primarily outline boundaries, without addressing the assessment of maturity.
Barmpoutis et al. \cite{48} used a segmentation model with dilated convolutions for binary segmentation of TLS and refined boundaries using an active contour model.
Wang et al. \cite{49} introduced a CNN-based model for segmenting TLS boundaries from WSI to compute prognostic biomarkers.
Chen et al. \cite{57} proposed a segmentation model that simultaneously segments TLS, lymphocytes, and tissue foreground for prognostic analysis in various cancer types. 
Van et al. \cite{multi_sacle_cnn_04_pixel_level} utilized a multi-resolution model to segment TLS from low-resolution images, capturing both coarse-grained and short-range contextual information.
In 2024, Van et al. \cite{multi_sacle_cnn_05_pixel_level} employed three datasets to perform binary segmentation of TLS.
Given the different values of three levels of maturity, Li et al. \cite{46} classified TLS into 1 of 3 grades based on the feature of the lymphocyte density map.
Unlike the works, we propose an end-to-end TLS segmentation model to achieve segmentation of TLS across three maturity stages, defining this task as TLS semantic segmentation.

\subsection{Contextual Learning for Segmentation in WSI}

Considering WSI with pyramidal resolutions, researchers typically use networks with low-resolution branch to capture the context of the target patch. Gu et al. \cite{multi_sacle_cnn_01_pixel_level} first introduced a low-resolution channel to the U-Net \cite{50} to encode contextual information, guiding the encoding and decoding processes of the network.
Ho et al. \cite{multi_sacle_cnn_02_pixel_level} developed a deep multi-resolution network with multiple encoder-decoder branches to extract more comprehensive contextual information. 
To ensure pixel-level spatial alignment of detail and context of the target patch,
Schmitz et al. \cite{multi_sacle_cnn_03_pixel_level} integrated CNN segmentation networks of different scales to incorporate contextual information across various resolutions for the segmentation task.
Van et al. \cite{multi_sacle_cnn_04_pixel_level} aligned feature maps at the same resolution using the \textit{Hooking} mechanism.
While the methods mentioned above emphasize the importance of contextual information, more effective approaches for learning and integrating contextual information are still being explored.


\begin{figure*}[ht]
  \centering
\includegraphics[width=0.9\textwidth]{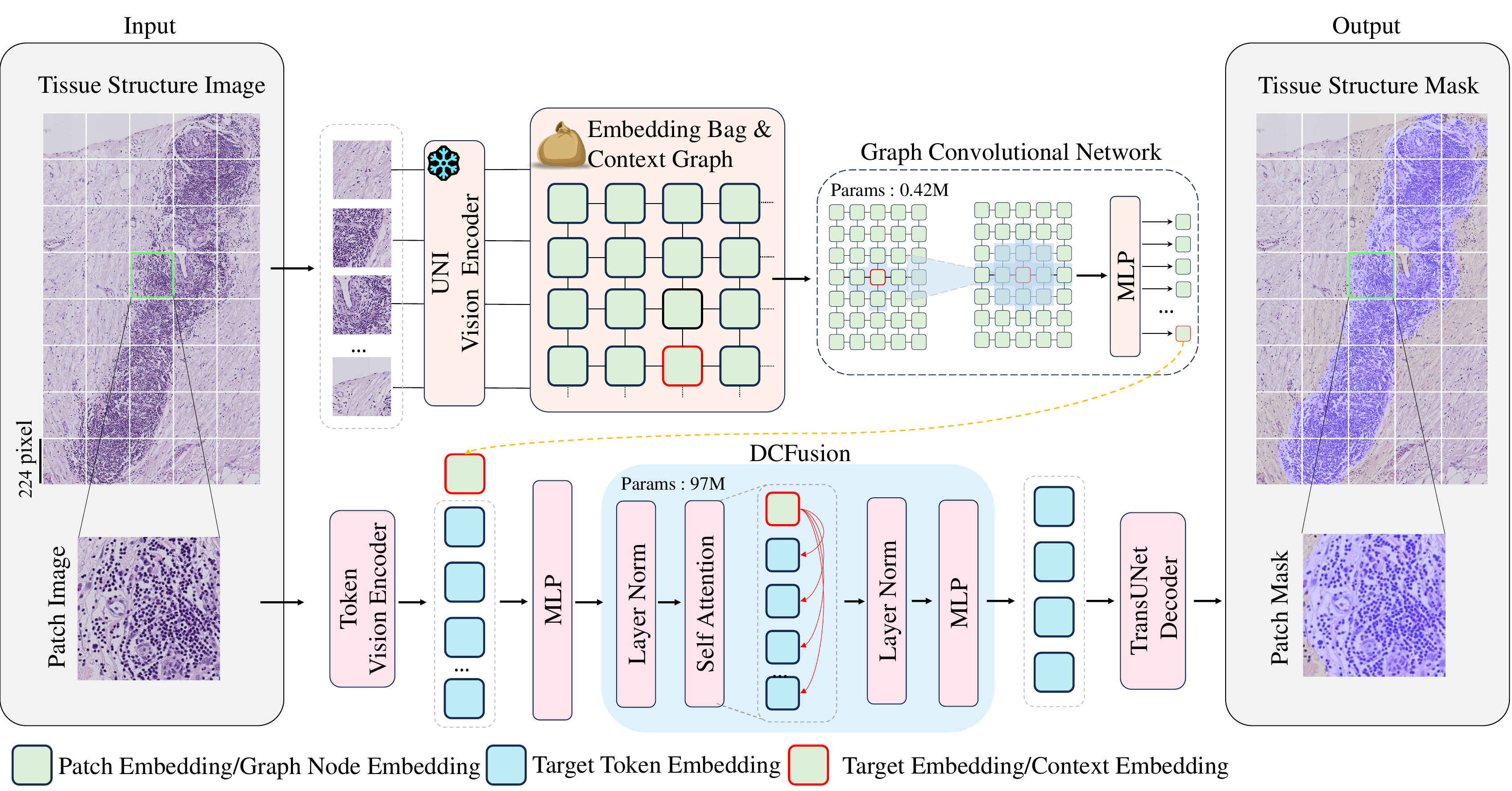} 
  \caption{The architecture of the proposed GCUNet. }
  \label{fig:method_frame}
\end{figure*}

\subsection{GNN-based Contextual Learning in WSI}

In Whole Slide Image analysis, Graph Neural Networks (GNNs) are commonly used to model contextual information between patches or cells, enabling the performance of tasks at either the WSI level or patch level.
Lu et al. \cite{contextual_learning_cell_HER2_pre_2022_MIA} constructed a cell graph to capture global contextual information for biomarker prediction in breast cancer. 
Richard et al. \cite{contextual_gnn_patchgcn_2021_miccai} used GCN to learn the global context of WSI for survival prediction.
Hou et al. \cite{multi_sacle_graph_01_wsi_level} introduced a heterogeneous graph to learn multi-scale contextual information for tumor typing and staging.
Shi et al. \cite{multi_sacle_graph_09_wsi_level} investigated cross-scale spatial context based on hierarchical graph for pathological primary tumor staging.
Li et al. \cite {31}  used dynamic graphs to describe the flexible interaction between patches in WSI to tumor typing and staging.
In addition to WSI-level tasks, GNN-based contextual learning has also been applied to WSI classification tasks in patch-level \cite{38, multi_sacle_graph_08_patch_level}. 
Compared with them, our model leverages GNNs to capture contextual information outside the target patch for the pixel-level task.




\section{Method}
\label{sec:method}
\subsection{Pipeline Overview}
Semantic segmentation of TLS aims to delineate the boundaries of TLS at different maturation stages (E-TLS, PET-TLS, and SEL-TLS) from WSI.
In this process, a TLS may be divided into multiple patches, some of which contain significant discriminative information (e.g., GC) that determines the maturation level of the TLS.
Therefore, given an image patch (target) to be segmented, the model needs to be aware of the contextual information outside target to discover the significant discriminative features. 
Our method consists of two steps: First, the multi-layer GCN iteratively aggregates long-range and fine-grained contextual information outside target. 
Then, DCFusion integrates the context and detail of target at the semantic level to predict the segmentation mask.
The proposed GCUNet architecture is illustrated in \Cref{fig:method_frame}.
\subsection{Context Graph Construction}
To model the contextual relationships of all patches in WSI, we construct a context graph \( \boldsymbol{G} = (\boldsymbol{V}, \boldsymbol{E}) \)  where \( \boldsymbol{V} \) denotes the set of patch features, and \( \boldsymbol{E} \) represents the set of edges that connect patches.
To be specific, the foreground of WSI is filtered following \cite{55}, and divided into non-overlapping patches, resulting in a set of \( N \) image patches \( \boldsymbol{P} = \{ \boldsymbol{p}_i \mid i=1 \ldots N \} \).
We use UNI \cite{56} to encode each \( \boldsymbol{p}_i \) into a feature vector \( \boldsymbol{v}_i \in \mathbb{R}^{1024} \). 
UNI is a transformer-based vision encoder for CPath, which has been pre-trained on millions of pathology images using a self-supervised method. 
Consequently, a WSI can be represented as a set of \( N \) nodes \( \boldsymbol{V} = \{ \boldsymbol{v}_i \mid i=1 \ldots N \} \).
Next, the undirected edge set \( \boldsymbol{E} = \{ \boldsymbol{v}_i \boldsymbol{v}_j \mid (i,j) \in \mathcal{H} \} \)  for the graph is determined based on the spatial connectivity of patches, where \( \mathcal{H} \) represents the set of naturally connected nodes using 4-connectivity.
\begin{table*}[ht]
\centering
\begin{tabular}{l l c c c c c c}
\hline
TASK & Data         & WSI & E-TLS & PET-TLS & SEL-TLS & NSEL-TLS & Total \\ \hline
\multirow{2}{*}{Seg4} & INHOUSE-PAAD & 108 & 2339 & 1586 & 611 & --      & 4536 \\
                      & TCGA-COAD    & 225 & 3496 & 1034 & 511 & --      & 5041 \\ \hline
\multirow{2}{*}{Seg3} & TCGA-BLCA    & 342 & --   & --   & 498 & 2538    & 3036 \\
                      & TCGA-LUSC    & 259 & --   & --   & 511 & 6688    & 7199 \\ \hline
\end{tabular}
\caption{Overview of Dataset Count. Seg4 involves semantic segmentation into four categories: background (BG), E-TLS, PET-TLS, and SEL-TLS. Seg3 divides the semantic segmentation task into three categories: BG, SEL-TLS, and NSEL-TLS.}
\label{tab:dataset}
\end{table*}
\subsection{Contextual Information Aggregation}
The context graph models the features and contextual relationships of each patch.
The adjacency matrix \( A = [a_{ij}]_{n \times n} \) is derived from the connection relationships between the graph nodes.  The elements of the adjacency matrix are defined as follows:
\[
a_{ij} = \begin{cases} 
1, & \text{if } [\boldsymbol{v}_i, \boldsymbol{v}_j] \in \boldsymbol{E} \\
0, & otherwise
\end{cases},
\tag{1}
\] 
where 
the feature matrix \( \boldsymbol{X}^{(0)} = \{ \boldsymbol{x}_1^{(0)}, \boldsymbol{x}_2^{(0)}, \dots, \boldsymbol{x}_N^{(0)} \} \in \mathbb{R}^{N \times 1024} \) represents the initial feature map for the \( N \) nodes, with each \( \boldsymbol{x}_i^{(0)} = \boldsymbol{v}_i \).
The aggregation of contextual information outside the patches over $t$ steps can be expressed as:
\[
\boldsymbol{X}^{(t)} = \text{F}_{\text{GCN}}(\boldsymbol{X}^{(t-1)}) = \sigma(\tilde{\boldsymbol{A}} \boldsymbol{X}^{(t-1)} \boldsymbol{W}^{(t-1)}), \tag{2}
\]
where \( \tilde{\boldsymbol{A}} = \boldsymbol{D}^{-1/2}(\boldsymbol{A} + \boldsymbol{I})\boldsymbol{D}^{-1/2} \) represents the normalized adjacency matrix, which is computed to balance the number of neighbors for each node, and \( \boldsymbol{D} \) denotes the degree matrix. 
The target node \( x_i \) aggregates features from its neighbors, progressively expanding the scope of its contextual information.

After \( T_0 \) aggregation steps, the feature of node \( i \) is updated from \( \boldsymbol{x}_i^{(0)} \) to \( \boldsymbol{x}_i^{(T_0)} \), incorporating contextual information from increasingly distant neighbors.
The set of neighbors within \( t \) steps from $i$-th node
is denoted as \( \text{Nera}_t(\boldsymbol{x}_i) = \{ \boldsymbol{x}_j \mid d(i,j) = t \} \), where \( d(i,j) \) represents the shortest path length between $i$-th node
and $j$-th node.
Therefore, the feature of the $i$-th node 
is updated based on the union of features from all neighbors up to \( T_0 \) steps, which can be expressed as:
\[
\boldsymbol{x}_i^{(T_0)} = \text{F}_{\text{GCN}}^*(\bigcup_{t=1}^{T_0} \text{Nera}_t(\boldsymbol{x}_i^{(0)})),  \tag{3}
\]
where, \( \text{F}_{\text{GCN}}^* \) is the graph convolution operation that updates the feature of the node \( i \) by aggregating information from its neighbors at increasing hops. 
This process enables the feature of the node to capture long-range contextual information, ultimately learning a richer representation of the target patch. Therefore, multiple GCN aggregation steps enable $\boldsymbol{v}_i$ to learn increasingly distant contextual information.

\subsection{Fusion of Detail and Contextual Information}

As illustrated in \Cref{fig:method_frame}, after obtaining the long-range contextual features \( \boldsymbol{z}_i^c = \boldsymbol{x}_i^{(T_0)} \in \mathbb{R}^{1 \times L} \) for the image patch \( \boldsymbol{p}_i \in \mathbb{R}^{H \times W \times 3} \), we utilize the encoder from TransUNet \cite{50} to extract the detailed features \( \boldsymbol{z}_i^d \in \mathbb{R}^{b^2 \times L} \), where \( b^2 = \frac{HW}{l^2} \) represents the number of tokens for the image patch \( p_i \).

Before the detailed features \( \boldsymbol{z}_i^d \) and the long-range contextual features \( \boldsymbol{z}_i^c \) are fed into the DCFusion module for fusion, the positional encoding is applied to the detailed features \( \boldsymbol{z}_i^d \). The two types of features are then concatenated into \( \boldsymbol{z}_i^{(0)} = [\boldsymbol{z}_i^c + \boldsymbol{e}_{\text{pos}}; \boldsymbol{z}_i^d] \in \mathbb{R}^{(b^2 + 1) \times L} \). The distinctiveness of the overall context in the detailed features \( \boldsymbol{z}_i^d \) is enhanced by the addition of positional encoding. 

DCFusion consists of \( \ell \) layers of multi-head attention (MSA) and a fully connected block. The final fused feature is computed as follows:
\begin{equation}
\begin{aligned}
\boldsymbol{z}_i^{(\ell-1)} &= \text{MSA}(\text{LayerNorm}(\boldsymbol{z}_i^{(\ell-2)})) + \boldsymbol{z}_i^{(\ell-2)} \\
\boldsymbol{z}_i^{(\ell)} &= \text{MSA}(\text{LayerNorm}(\boldsymbol{z}_i^{(\ell-1)})) + \boldsymbol{z}_i^{(\ell-1)}
\end{aligned}
,
\tag{5}
\end{equation}
where \( \boldsymbol{z}_i^{(*)} \) represents the output of the attention hidden layers, and \( \boldsymbol{z}_i^{(\ell)} \in \mathbb{R}^{(b^2 + 1) \times L} \) is the final output of the semantic feature fusion. MLP refers to a learnable fully connected layer.

\subsection{Decoding Fused Features for Segmentation}
After obtaining the fused features \( \boldsymbol{z}_i^{(\ell)} \) that integrate both detailed and contextual information of \( p_i \), we assume that the token features and contextual information within \( \boldsymbol{z}_i^{(\ell)} \) have been fully integrated. Therefore, we only select the features corresponding to the token positions, denoted as \( \boldsymbol{z}_i^{\prime (\ell)} \in \mathbb{R}^{b^2 \times L} \). These features are then fed into the decoder of TransUNet \cite{compared_attention_transunet} to predict the segmentation mask \( \boldsymbol{y}_m^{\text{pred}} \in \mathbb{R}^{k \times H \times W} \), where $k$ refers to the predefined number of categories for the segmentation targets.
Finally, the segmentation loss is computed using cross-entropy, and the network is optimized through backpropagation:
\[
L_{\text{bce}} = -\frac{1}{H \times W} \sum_{h,w} \left[ \boldsymbol{y}_m^{\text{target}}(h, w) \log(\boldsymbol{y}_m^{\text{pred}}(h, w)) \right]. \tag{6}
\]
GCUNet is trained in an end-to-end manner.

\begin{table*}[t]
\centering
\begin{tabular}{cccccccccc}  
\toprule
\multirow{2}{*}{Type} & \multirow{2}{*}{Method} & \multicolumn{4}{c}{INHOUSE-PAAD} & \multicolumn{4}{c}{TCGA-COAD} \\ 
\cmidrule{3-10} 
                      &                         & mF1   & mIOU  & mP    & mR    & mF1   & mIOU  & mP    & mR    \\ 
\midrule
\multirow{2}{*}{CNN}   & U-Net \cite{50}                 & 0.559 & 0.418 & 0.559 & 0.563 & 0.556 & 0.423 & 0.557 & 0.555 \\ 
                      & Attention-UNet \cite{compared_attention_unet}         & 0.536 & 0.399 & 0.550 & 0.536 & 0.528 & 0.398 & 0.547 & 0.547 \\ 
\midrule
\multirow{4}{*}{Trans} & Swin-UNet \cite{compared_attention_swinunet}              & 0.558 & 0.416 & 0.562 & 0.569 & 0.556 & 0.424 & 0.567 & 0.551 \\ 
                      & H2Former \cite{compared_h2former}               & 0.543 & 0.394 & 0.542 & 0.525 & 0.523 & 0.429 & 0.429 & 0.431 \\ 
                      & DTMFormer \cite{compared_dtmformer}              & 0.531 & 0.394 & 0.539 & 0.528 & 0.543 & 0.413 & 0.544 & 0.543 \\ 
                      & Transunet (baseline) \cite{compared_attention_transunet}   & 0.555 & 0.415 & 0.555 & 0.556 & 0.560 & 0.427 & 0.575 & 0.559 \\ 
\midrule
Multi-res              & HookNet \cite{multi_sacle_cnn_04_pixel_level}                & 0.580 & 0.427 & 0.608 & 0.570 & 0.597 & 0.463 & 0.603 & 0.593 \\ 
\midrule
\multirow{1}{*}{GNN} & GCUNet (ours)            & \textbf{0.623} & \textbf{0.474} & \textbf{0.655} & \textbf{0.613} & \textbf{0.665} & \textbf{0.523} & \textbf{0.676} & \textbf{0.660} \\ 
\bottomrule
\end{tabular}
\caption{The performance of our method and the SOTA on the Seg4 task using the INHOUSE-PAAD and TCGA-COAD datasets.}
\label{tab:1}
\end{table*}
\section{Experiments}
\label{sec:experi}
\subsection{Datasets}
For the pancreatic adenocarcinoma (INHOUSE-PAAD) dataset, we select two adjacent tissue sections from each patient:
one for H\&E staining and the other for mIHC staining, including CD3, CD20, CD21, and CD23).
Aided by mIHC, pathologists annotated the boundaries and classified them into three maturation stages in WSI.
For The Genome Cancer Atlas (TCGA)  colon adenocarcinoma (TCGA-COAD) dataset , the H\&E data were downloaded from TCGA and cleaned by excluding low-quality WSI, such as those containing artifacts, folds, or large areas of necrosis. 
TLSs at three maturity levels were annotated by pathologists without mIHC assistance.
For the bladder urothelial carcinoma (TCGA-BLCA) and lung squamous cell carcinoma (TCGA-LUSC) datasets, we use public annotations \cite{multi_sacle_cnn_05_pixel_level} that highlight GC and TLS, without distinguishing between maturation stages of the TLS.
To prepare these data for TLS semantic segmentation task, we first exclude WSIs that did not contain TLS and classifiy the TLS into SEL-TLS and Non-SEL-TLS (NSEL-TLS) based on the presence of GC. 
NSEL-TLS does not contain GC and cannot be distinguished as either E-TLS or PET-TLS.
Therefore, we define TLS in the TCGA-BLCA and TCGA-LUSC datasets as two categories: SEL-TLS and NSEL-TLS.
Ultimately, the TLS semantic segmentation datasets for four types of cancer were collected.
There are two tasks for the four datasets: four-class semantic segmentation (Seg4) and three-class semantic segmentation (Seg3).
The INHOUSE-PAAD and TCGA-COAD datasets were utilized for the \textbf{Seg4}, entailing semantic segmentation into four categories: background (\textbf{BG}),\textbf{ E-TLS}, \textbf{PET-TLS}, and \textbf{SEL-TLS}.
 The TCGA-BLCA and TCGA-LUSC datasets were utilized for \textbf{Seg3}, with categories designated as \textbf{BG}, \textbf{NSEL-TLS}, and \textbf{SEL-TLS}. 
 For each dataset, we randomly divide the data into training, validation, and testing sets in a ratio of 6:2:2.


\begin{table*}[t]
\centering
\begin{tabular}{cccccccccc}  
\toprule
\multirow{2}{*}{Type} & \multirow{2}{*}{Method} & \multicolumn{4}{c}{TCGA-BLCA} & \multicolumn{4}{c}{TCGA-LUSC} \\ 
\cmidrule{3-10} 
                      &                         & mF1   & mIOU  & mP    & mR    & mF1   & mIOU  & mP    & mR    \\ 
\midrule
\multirow{2}{*}{CNN}   & U-Net \cite{50}                  & 0.620  & 0.486 & 0.617 & 0.626 & 0.546 & 0.472 & 0.524 & 0.572 \\ 
                      & Attention-UNet \cite{compared_attention_unet}         & 0.590  & 0.468 & 0.616 & 0.592 & 0.535 & 0.458 & 0.514 & 0.560 \\ 
\midrule
\multirow{4}{*}{Trans} & Swin-UNet \cite{compared_attention_swinunet}              & 0.620  & 0.484 & 0.628 & 0.626 & 0.548 & 0.472 & 0.614 & 0.564 \\ 
                      & H2Former \cite{compared_h2former}               & 0.621 & 0.488 & 0.619 & 0.624 & 0.566 & 0.473 & 0.579 & 0.568 \\ 
                      & DTMFormer \cite{compared_dtmformer}              & 0.566 & 0.440 & 0.562 & 0.572 & 0.556 & 0.457 & 0.555 & 0.558 \\ 
                      & TransUNet (baseline) \cite{compared_attention_transunet}   & 0.608 & 0.480 & 0.618 & 0.602 & 0.545 & 0.471 & 0.522 & 0.575 \\ 
\midrule
Multi-res              & HookNet \cite{multi_sacle_cnn_04_pixel_level}                & 0.629 & 0.500 & 0.666 & 0.608 & 0.549 & 0.476 & 0.552 & 0.570 \\ 
\midrule
\multirow{1}{*}{GNN} & GCUNet (ours)            & \textbf{0.740}  & \textbf{0.602} & \textbf{0.744} & \textbf{0.744} & \textbf{0.703} & \textbf{0.576} & \textbf{0.705} & \textbf{0.702} \\ 
\bottomrule
\end{tabular}
\caption{Comparison of segmentation results for Our model with other models on the Seg3 task using the TCGA-BLCA and TCGA-LUSC datasets.}
\label{tab:2}
\end{table*}

\subsection{Implementation details}
To evaluate the effectiveness of the proposed GCUNet, we compare it with several models based on CNN, Transformer, and multi-resolution approaches, including U-Net \cite{50}, Attention U-Net \cite{compared_attention_transunet}, SwinUNet \cite{compared_attention_swinunet}, TransUNet \cite{compared_attention_transunet}, H2Former \cite{compared_h2former}, DTMFormer \cite{compared_dtmformer}, and HookNet \cite{multi_sacle_cnn_04_pixel_level}. TransUNet was used as the baseline.

We apply OTSU \cite{55} to distinguish the foreground region. 
In most experiments, the patch size is set to 224$\times$224 with a spatial resolution of 1 $\mu$m/pixel. 
For HookNet \cite{multi_sacle_cnn_04_pixel_level}, the image patch size was set to 256$\times$256 to ensure full alignment of feature maps with different resolutions.
Softmax is employed in the aggregation function of the GCN layer, with the the temperature constant initialized as a learnable parameter set to 1.
The hidden feature dimension is set to 128.
We use an encoder \cite{multi_sacle_cnn_04_pixel_level} to extract patch details and employ a 12-layer attention network, where each layer consists of 12 attention heads and has a hidden feature dimension of 768.
During training, the patch size is set to 16, and the learning rate is set to $5\times10^{-5}$. 
In the experiment, We report the F1 score, IoU, Precision, and Recall for each category.
The average values of these metrics across categories, denoted as mF1, mIoU, mP, and mR, are used to evaluate overall segmentation performance.

\begin{table*}[t]
\centering
\begin{tabular}{cccccccccc}  
\toprule
\multirow{2}{*}{Type} & \multirow{2}{*}{Method} & \multicolumn{4}{c}{INHOUSE-PAAD} & \multicolumn{4}{c}{TCGA-COAD} \\ 
\cmidrule{3-10} 
                      &                         & BG    & E-TLS  & PET-TLS  & SEL-TLS  & BG    & E-TLS  & PET-TLS  & SEL-TLS  \\ 
\midrule
\multirow{2}{*}{CNN}   & U-Net \cite{50}                  & 0.889 & 0.444 & 0.409 & 0.496 & 0.930 & 0.426 & 0.415 & 0.451 \\ 
                      & Attention-UNet \cite{compared_attention_unet}         & 0.889 & 0.348 & 0.472 & 0.433 & 0.913 & 0.429 & 0.350 & 0.419 \\ 
\midrule
\multirow{4}{*}{Trans} & Swin-UNet \cite{compared_attention_swinunet}              & 0.882 & 0.455 & 0.381 & 0.514 & 0.929 & 0.437 & 0.443 & 0.417 \\ 
                      & H2Former \cite{compared_h2former}               & 0.888 & 0.370 & 0.406 & 0.458 & 0.929 & 0.429 & 0.431 & 0.383 \\ 
                      & DTMFormer \cite{compared_dtmformer}               & 0.886 & 0.401 & 0.425 & 0.413 & 0.911 & 0.416 & 0.469 & 0.397 \\ 
                      & TransUNet (baseline) \cite{compared_attention_transunet}   & 0.888 & 0.418 & 0.408 & 0.506 & 0.930 & 0.452 & 0.461 & 0.339 \\ 
\midrule
Multi-res              & HookNet \cite{multi_sacle_cnn_04_pixel_level}                & 0.837 & 0.483 & 0.474 & 0.528 &\textbf{0.945} & 0.501 & 0.419 & 0.524 \\ 
\midrule
\multirow{1}{*}{GNN} & GCUNet (ours)            & \textbf{0.894} & \textbf{0.493} & \textbf{0.548} & \textbf{0.555} & 0.933 & \textbf{0.564} & \textbf{0.544} & \textbf{0.621} \\ 
\bottomrule
\end{tabular}
\caption{Comparison of F1 for GCUNet and other models on the INHOUSE-PAAD and TCGA-COAD datasets.}
\label{tab:seg4_comparison}
\end{table*}

\subsection{Comparisons with State-of-the-Art Methods}
\Cref{tab:1} presents a performance comparison between GCUNet and other methods on the Seg4. GCUNet significantly outperforms other methods across all four evaluation metrics. Compared to baseline, GCUNet improves mF1 by 0.068 and 0.105, representing an increase of 11.72\% and 18.75\% on INHOUSE-PAAD and TCGA-COAD, respectively.
Notably, the multi-resolution network HookNet exhibits suboptimal performance. HookNet improves mF1 by 0.021 and 0.041 compared to the U-Net and outperforms the Trans series methods on two datasets.
These results highlight the importance of leveraging contextual information outside target patches in TLS semantic segmentation.
At the same time, GCUNet achieves a 7.41\% improvement in mF1 over HookNet. This performance is attributed to the utilization of long-range and fine-grained contextual information.
On the TCGA-COAD dataset, GCUNet also outperforms the second-best model HookNet, with improvements of 11.39\% in mF1 and 12.96\% in mIOU.
GCUNet significantly outperforms all other methods in the Seg4, confirming the effectiveness of the proposed model.

\Cref{tab:2} displays the comparative performance of GCUNet against other models for the Seg3 task.
The Seg3 task involves three categories: BG, SEL-TLS, and NSEL-TLS. 
Compared to Seg4, Seg3 is less challenging, resulting in enhanced overall performance for all the methods evaluated.
GCUNet also outperforms the best on the TCGA-BLCA and TCGA-LUSC datasets, achieving mF1 scores of 0.74 and 0.703, respectively.
GCUNet improves mF1 by 0.132 compared to baseline.
Meanwhile, GCUNet achieves the best performance on the TCGA-LUSC dataset, where mF1 and mIOU are improved by 0.158 and 0.105 over the baseline, respectively.
Significantly, HookNet retains its position as the second-best model on the TCGA-BLCA dataset. 
However, its lead has been considerably reduced, with an mF1 score only marginally higher by 0.009 compared to U-Net.
On the TCGA-LUSC dataset, the second-best model is H2Former.
The experiments indicate the advantage of HookNet decreases in TLS semantic segmentation with fewer categories. 
\begin{figure*}[htbp]
  \centering
  \includegraphics[width=0.9\textwidth]{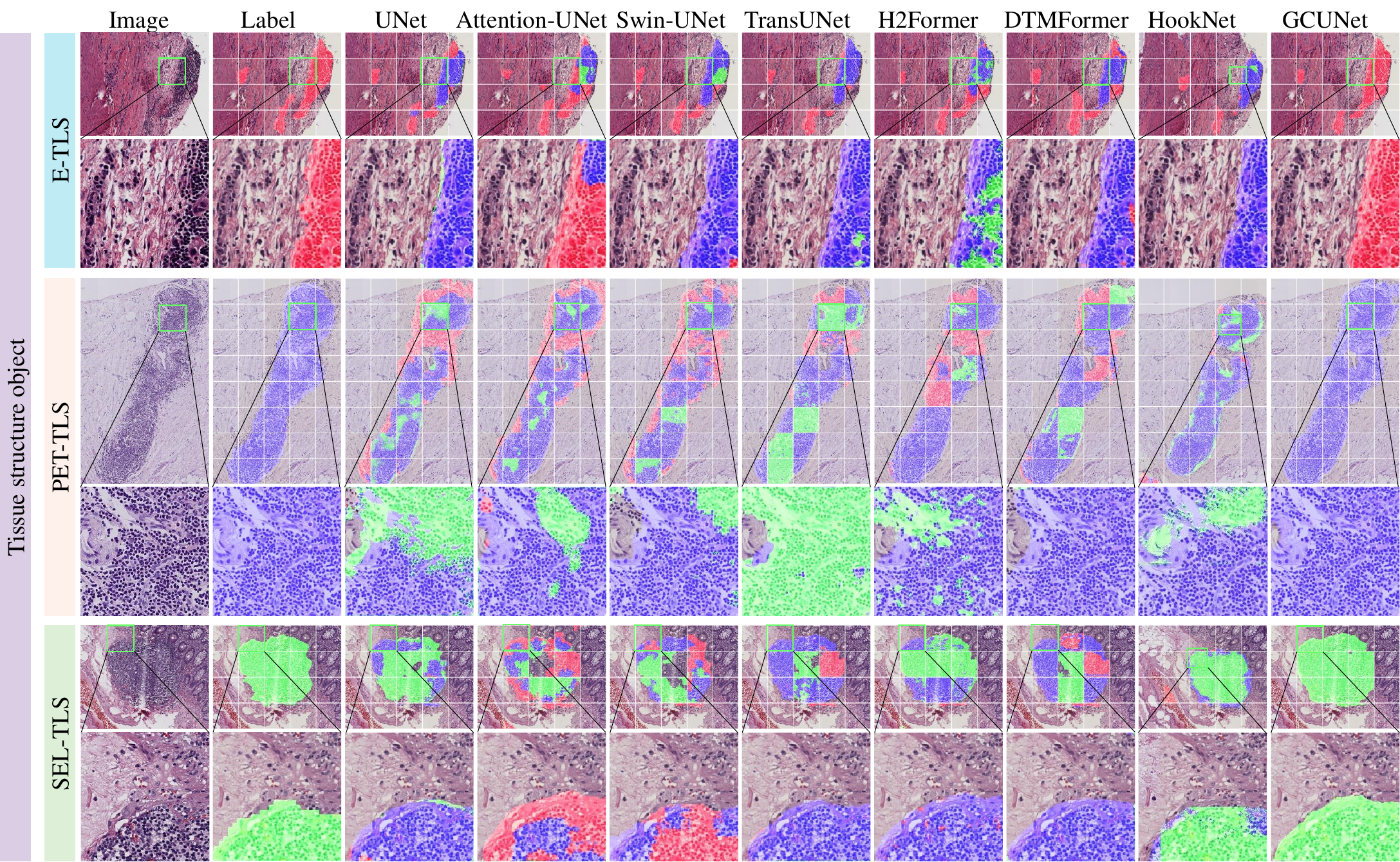} 
  \caption{Visualization of segmentation results for three types of TLS—E-TLS, PET-TLS, and SEL-TLS.
E-TLS are highlighted in red, PET-TLS in blue, and SEL-TLS in green according to our annotation guidelines. 
Each TLS contains a pair of images in two rows: the top row shows a global view, while the bottom row provides a detailed view of the highlighted region.}
  \label{fig:visu}
\end{figure*}
To analyze the factors for the performance improvement of GCUNet, we discuss the mF1 scores for BG, E-TLS, PET-TLS, and SEL-TLS in the Seg4.

As shown in the \Cref{tab:seg4_comparison}, the superior segmentation of the three TLS maturity by GCUNet. Our model can identify key features by capturing both fine-grained and long-range contextual information beyond the target patch.
HookNet lacks robustness in distinguishing the background, resulting in two extreme outcomes on INHOUSE-PAAD and TCGA-COAD.
This suggests that the performance of TLS semantic segmentation is hindered by contextual information outside the target. 
The improvement in model performance is primarily due to the ability to distinguish the three maturity levels and the long-range, fine-grained contextual information provided by GNNs.
\subsection{Visualisation}
\Cref{fig:visu} show that GCUNet achieves the most accurate segmentation, closely matching the ground truth labels, particularly in capturing fine-grained details and distinguishing between TLS maturity levels. 
Other models struggle to utilize discriminative features effectively due to their inability to fully leverage contextual information beyond the target patch, resulting in poor consistency in TLS segmentation.
GCUNet consistently performs well across all TLS types, especially in SEL-TLS regions containing germinal centers, where it produces clearer and more accurate boundaries than other methods.

\subsection{Ablation Studies}
Drawing on both quantitative and qualitative analysis, we further investigate several factors that influence model performance. These factors include the method of integrating detailed and contextual information, the range of required contextual information, and information granularity of the patches.

\begin{figure}
  \centering
  \includegraphics[width=0.9\linewidth]{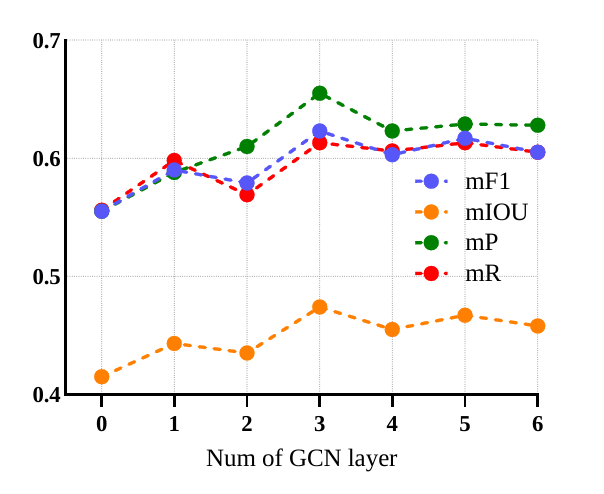}
  \caption{The impact of changing the number of GCN layers on four evaluation metrics.}
  \label{fig:5a}
\vspace{-10pt}  

\end{figure}

\textbf{Number of GCN layers}: \Cref{fig:5a} presents the results of an ablation experiment exploring the effect of varying the number of GCN aggregation layers in Seg4 on the INHOUSE-PAAD dataset. Let $\boldsymbol{N_c} = (0, 1, \dots, 6)$ represent the number of GCN layers, with the baseline corresponding to $\boldsymbol{N_c} = 0$. The number of GCN layers influences how far contextual information can be propagated for the target patch. 
We iteratively adjust $\boldsymbol{N_c}$ and compare the corresponding segmentation performance, keeping all other parameters constant. 
The figure shows that model performance is sensitive to the number of GCN layers. 
Segmentation performance significantly decreases as the distance of aggregated information is reduced, with the poorest performance observed when no contextual information is used.
However, as the distance of aggregated information increases further, the performance slightly declines and then stabilizes.
Through multiple experiments with varying numbers of layers, we find that the model achieves the best results when the number of GCN layers is set to 3.


\begin{figure}[ht]
  \centering
  \includegraphics[width=0.9\linewidth]{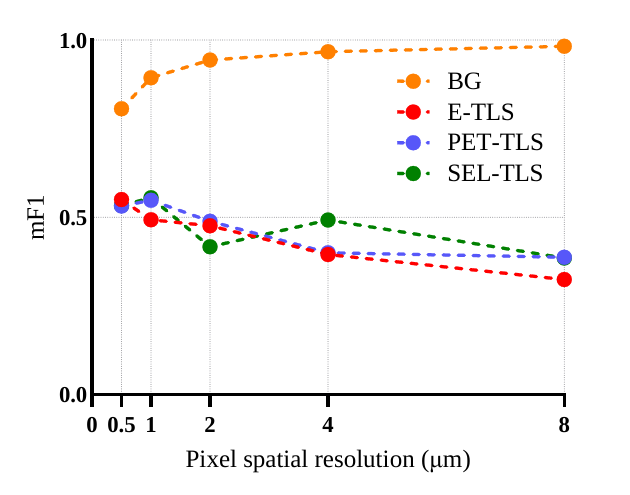}
  \caption{The impact of changing the pixel spatial resolution.}
  \label{fig:5b}
\end{figure}
\textbf{Information Granularity}:
\Cref{fig:5b} illustrates the impact of information granularity on the INHOUSE-PAAD dataset. We set the number of GCN layers to 3, the patch size to 224$\times$224, and the pixel spatial resolutions to $\boldsymbol{mpp} = (0.5, 1, 2, 4, 8)$. The table presents the mF1 scores of the segmentation results.
As the spatial resolution increases, the model becomes better at distinguishing the background, but the performance on TLS decreases significantly.
The model performs the worst on the background, although the distinction between the three TLS categories remains relatively balanced.
These experiments suggest that TLS semantic segmentation requires fine-grained information. 
At a spatial resolution of 1 $\mu$m/pixel, the model achieved relatively balanced performance across both background and TLS categories.
\begin{table}[ht]
\centering
\begin{tabular}{lcccccc}
\toprule
Method        & \multicolumn{2}{c}{INHOUSE-PAAD} & \multicolumn{2}{c}{TCGA-COAD} \\
\cmidrule{2-5}
              & mF1   & mIOU  & mF1   & mIOU  \\
\midrule
w/o-context   & 0.555 & 0.415 & 0.560 & 0.427 \\
Cat           & 0.586 & 0.442 & 0.651 & 0.508 \\
Dot           & 0.610 & 0.462 & 0.655 & 0.514 \\
DCFusion (ours)     & \textbf{0.623} & \textbf{0.474} & \textbf{0.665} & \textbf{0.523} \\
\bottomrule
\end{tabular}
\caption{The impact of contextual information and detailed information fusion methods on model performance in the INHOUSE-PAAD and TCGA-COAD datasets.}
\label{tab:context_fusion}
\end{table}

\textbf{Fuison strategy}:
After determining the optimal resolution and number of GCN layers, we conduct experiments on various fusion methods for integrating detailed and contextual information of the target, using a pixel resolution of 1.0 $\mu$m/px and 3 GCN layers, on the INHOUSE-PAAD and TCGA-COAD datasets.
We use several fusion methods, including: Without Contextual Information (w/o-context), which does not incorporate contextual information and serves as the baseline for comparison; Concatenation (Cat), which fuses target details and contextual information by concatenating them; Dot Product (Dot), which combines target details and contextual information using a dot product operation; and DCFusion, the fusion strategy of GCUNet, which employs a self-attention mechanism for semantic-level fusion of the two types of information.

\Cref{tab:context_fusion} demonstrate that all fusion strategies effectively utilize contextual information. 
Among the fusion strategies, Cat outperformed other basic strategies. 
DCFusion achieved the best performance on both datasets, achieving mF1 increased by 9.25\% and 17.3\% over baseline, respectively.
These results highlight the importance of semantic-level fusion of target detail and contextual information.

\section{Conclusion}
\label{sec:results}



In this paper, we introduced a new task of TLS semantic segmentation in WSI and proposed a GNN-based contextual learning network GCUNet. 
GCUNet used GCNs to flexibly aggregate long-range and fine-grained contextual information beyond the target patch, while the designed DCFusion performed semantic-level fusion of detailed and contextual information to predict patch masks.
We collected four TLS semantic segmentation datasets and released annotations for three of them (TCGA-COAD, TCGA-LUSC, and TCGA-BLCA), comprising 826 WSIs and 15,276 TLSs.
Our results on these datasets demonstrated the superiority of GCUNet.

{
    \small
    \bibliography{main}
}

\end{document}